\documentclass{bmvc2k}
\usepackage{amsfonts}
\usepackage{multirow}

\title{A Prototype Unit for Image De-raining \\ using Time-Lapse Data}

\addauthor{Jaehoon Cho}{jh.cho@hyundai.com}{1}
\addauthor{Minjung Yoo}{ymj2471@kau.kr}{2}
\addauthor{Jini Yang}{sheep0901@kau.kr}{2}
\addauthor{Sunok Kim}{sunok.kim@kau.ac.kr}{2}

\addinstitution{
 Hyundai Motor Company, \\
 Seoul, Korea
}
\addinstitution{
 Artificial Intelligence,\\
 Korea Aerospace University,\\
 Goyang, Korea
}

\runninghead{Cho et al.}{A Prototype Unit for Image De-raining using Time-Lapse Data}

\usepackage{booktabs}
\usepackage{pifont}
\newcommand{\cmark}{\ding{51}}

\begin{document}

\maketitle

\begin{abstract}

 We address the challenge of single-image de-raining, a task that involves recovering rain-free background information from a single rain image. While recent advancements have utilized real-world time-lapse data for training, enabling the estimation of consistent backgrounds and realistic rain streaks, these methods often suffer from computational and memory consumption, limiting their applicability in real-world scenarios. In this paper, we introduce a novel solution: the Rain Streak Prototype Unit (RsPU). The RsPU efficiently encodes rain streak-relevant features as real-time prototypes derived from time-lapse data, eliminating the need for excessive memory resources. Our de-raining network combines encoder-decoder networks with the RsPU, allowing us to learn and encapsulate diverse rain streak-relevant features as concise prototypes, employing an attention-based approach. To ensure the effectiveness of our approach, we propose a feature prototype loss encompassing cohesion and divergence components. This loss function captures both the compactness and diversity aspects of the prototypical rain streak features within the RsPU. Our method evaluates various de-raining benchmarks, accompanied by comprehensive ablation studies. We show that it can achieve competitive results in various rain images compared to state-of-the-art methods.

\end{abstract}

\section{Introduction}
\label{sec:intro}

Image de-raining is an important task in computer vision as the rain streaks hinder visibility and deteriorate the robustness of most outdoor vision systems.
It has been widely applied in many tasks such as object detection~\cite{li2019single,fu2019lightweight}, semantic segmentation~\cite{cho2020single,jiang2020multi}, autonomous driving~\cite{huang2021memory,guo2021efficientderain}, or surveillance system~\cite{li2021online,li2021comprehensive}, as an essential pre-processing step.

\begin{figure}[!t]
	\centering
	\subfigure[Input]
	{\includegraphics[width=0.195\textwidth]{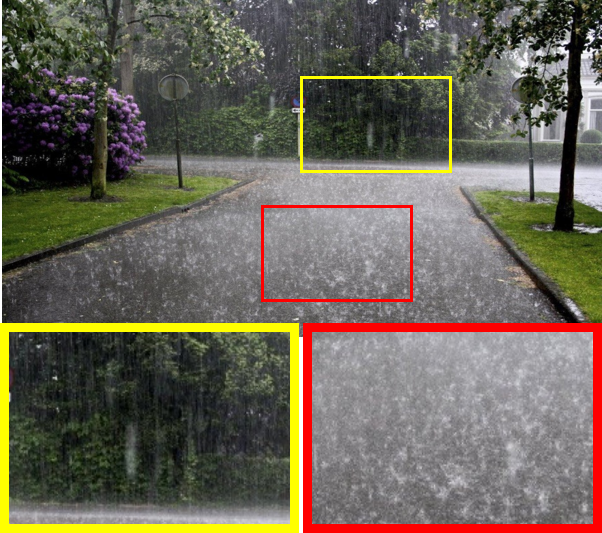}}
    \subfigure[SIRR~\cite{wei2019semi}]
	{\includegraphics[width=0.195\textwidth]{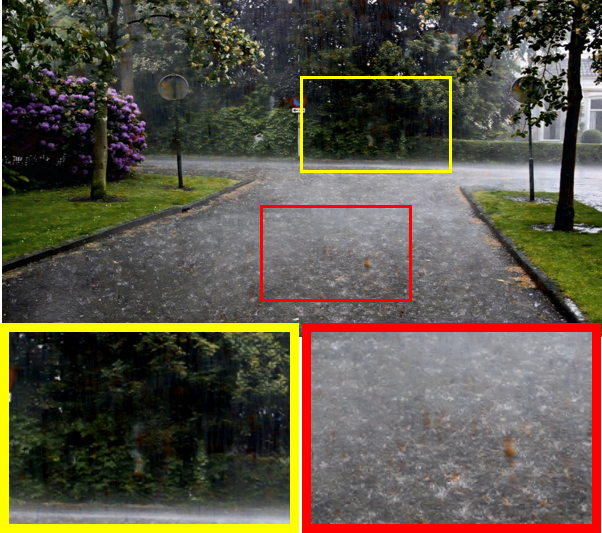}}
    \subfigure[MPRNet~\cite{zamir2021multi}]
	{\includegraphics[width=0.195\textwidth]{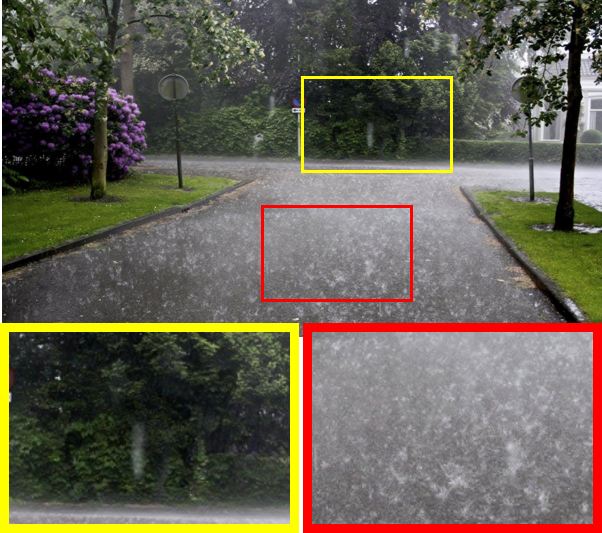}}
    \subfigure[Syn2Real~\cite{yasarla2020syn2real}]
	{\includegraphics[width=0.195\textwidth]{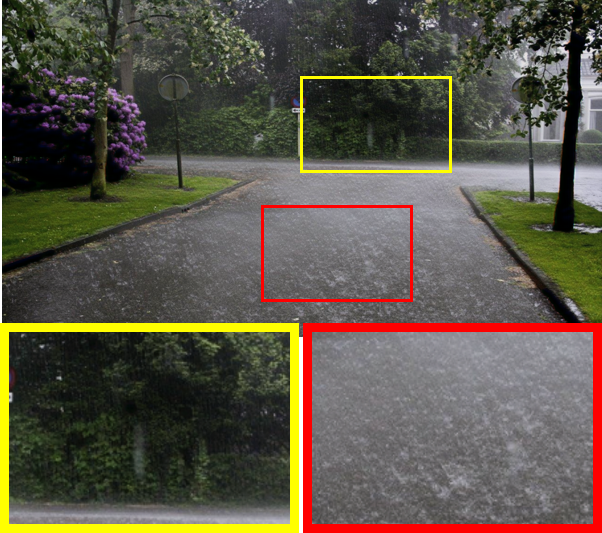}}
	\subfigure[RCDNet~\cite{wang2020model}]
	{\includegraphics[width=0.195\textwidth]{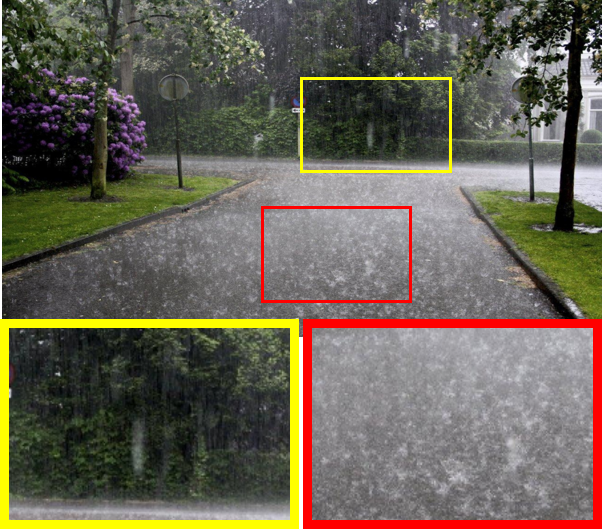}}\\
	\vspace{-8pt}
	\subfigure[PReNet~\cite{ren2019progressive}]
	{\includegraphics[width=0.195\textwidth]{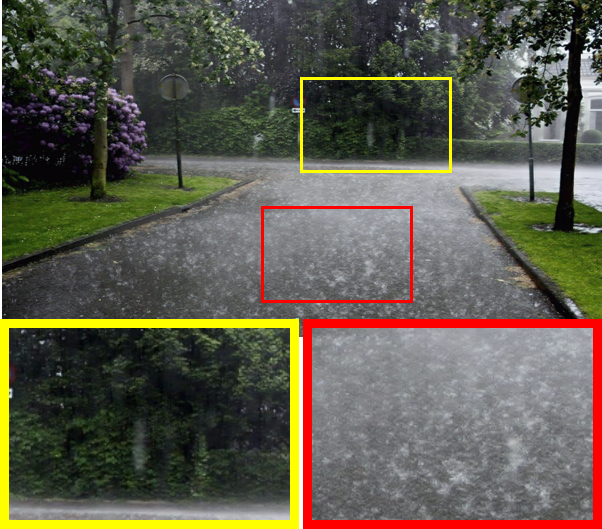}}
    \subfigure[Restormer~\cite{zamir2022restormer}]
	{\includegraphics[width=0.195\textwidth]{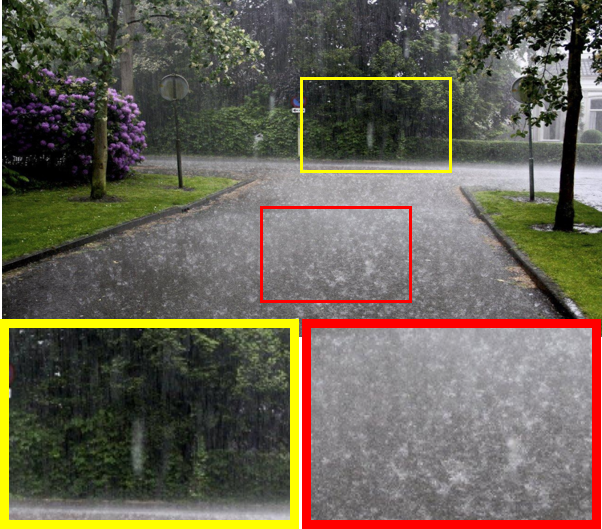}}
    \subfigure[MAXIM~\cite{tu2022maxim}]
	{\includegraphics[width=0.195\textwidth]{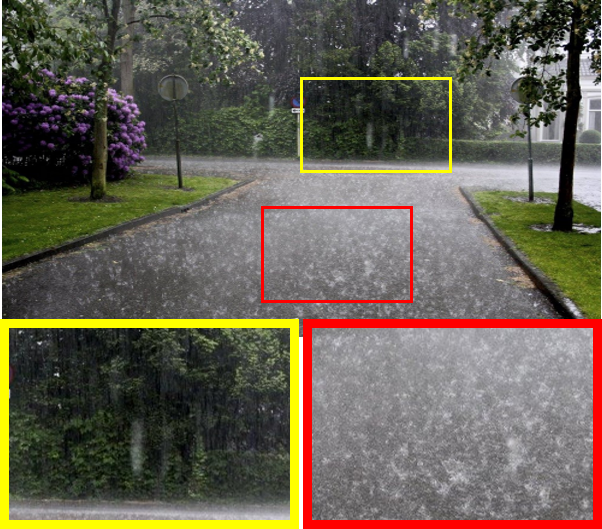}}
    \subfigure[DRSformer~\cite{chen2023learning}]
	{\includegraphics[width=0.195\textwidth]{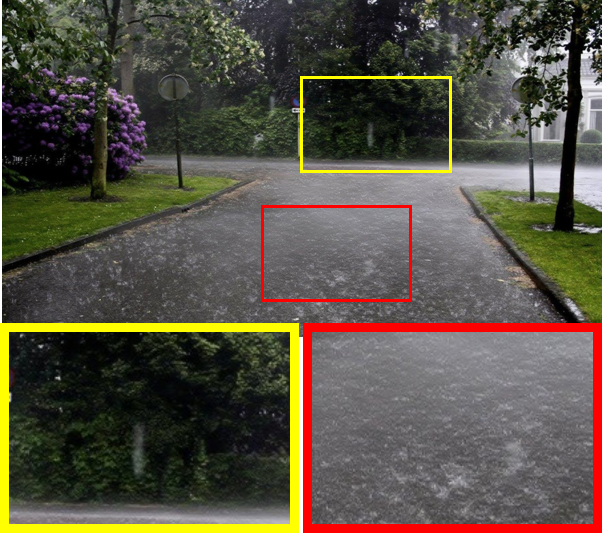}}
	\subfigure[Ours]
	{\includegraphics[width=0.195\textwidth]{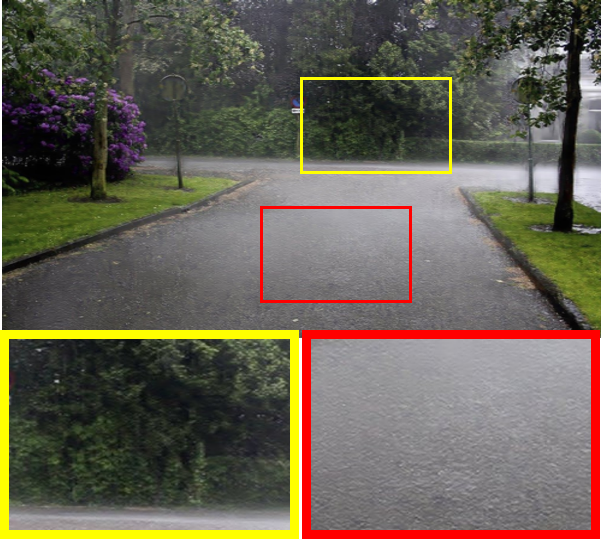}}\\
	\vspace{3pt}
	\caption{The de-rained results using the state-of-the-art de-rained method and the proposed method on a real rainy image.
Our method generates better-de-raining results than state-of-the-art methods on the real rainy image.}\label{fig:1} \vspace{-21pt}
\end{figure}

Many existing approaches to single-image-based de-raining employ deep Convolutional Neural Networks (CNNs) ~\cite{fu2017clearing,fu2017removing,yang2017deep,yang2019joint,li2018recurrent,yang2019scale,wang2020model}. Approaches utilizing transformers~\cite{chen2023learning,valanarasu2021transweather,zamir2022restormer,tu2022maxim} also have been employed for single-image-based de-raining and have shown a better improvement.
These methods are trained in a supervised manner using synthetic datasets containing both rain and rain-free images. 
However, the dissimilarities between synthetic and real rain images are apparent. As shown in Fig.~\ref{fig:1}, real rain streaks exhibit complex and diverse patterns in terms of size, scale, direction, and density. Consequently, methods trained on synthetic data struggle to handle real rain images effectively~\cite{yang2020single,huang2021memory,wang2019spatial}. Semi-supervised methods~\cite{wei2019semi,yasarla2020syn2real,huang2021memory} use synthetic and unpaired real rain data for training, yet they remain mainly reliant on synthetic data, leading to performance degradation when applied to real rain images.

To address the requirement for rain-clean image pairs, SPANet~\cite{wang2019spatial} and TimeLapsNet~\cite{cho2020single} introduced real rain datasets characterized by consistent backgrounds and varying rain streaks. TimeLapsNet~\cite{cho2020single} focused on estimating unchanging backgrounds while accommodating temporal changing rain streaks. Recently, MemoryNet~\cite{cho2022memory} leveraged an external memory network to capture rain streak features across real rain datasets. Especially, they showed remarkable results on real rainy images.  As a real-world application, single-image-based de-raining often needs to be executed on devices with limited computing power and memory in practice. However, heavy parameters from ~\cite{cho2020single}  and external memory networks from ~\cite{cho2022memory} hinder practical applicability.

Building on leveraging real rain datasets, this paper proposes a novel learning framework centered around a memory-efficient representation of rain streak features, named the Rain-streak Prototype Unit (RsPU). By encoding prototypical rain streak features using an attention mechanism, the RsPU effectively captures the essence of rain streaks. The process involves applying an attention operator on the encoder's encoded features, assigning rain streak weights to pixels, and generating a rain streak attention map, shaping the prototypical rain streak features. These prototypes are formed by aggregating local encoding vectors guided by rain streak weights. Multiple attention operators are employed to generate diverse prototype candidates. We propose a feature prototype loss that combines cohesion and divergence components to enhance the distinctiveness of the prototypical rain streak features. Cohesion loss minimizes intra-class variability to facilitate the clustering similar rain streaks, while divergence loss enforces diversity among prototypes.

We summarize our contributions as follows:  i) the introduction of the Rain-streak Prototype Unit (RsPU) for encoding diverse rain-streak features as prototypes without additional memory consumption, ii) the proposal of a feature prototype loss to enhance the discriminative capabilities of prototypical rain-streak features, and iii) the demonstration of state-of-the-art performance across various de-raining benchmarks, along with the adaptability of our method in real-world scenarios.
\vspace{-13pt}

\section{Related Work}
\label{sec:related}
\vspace{-5pt}
\subsection{Single Image De-raining}
Since obtaining the paired real dataset is challenging, most methods have used paired synthetic datasets. Fu \textit{et al.},~\cite{fu2017clearing} firstly proposed a deep learning-based method with multi-layer CNN to extract and remove the rain streaks and presented the deep detail network (DDN)~\cite{fu2017removing} learning a mapping function from a rainy image to the clean image. 
JORDER~\cite{yang2017deep} attempted to jointly handle the image de-raining with detecting the rain region and extended their work with the contextual dilated networks (JORDER-E~\cite{yang2019joint}).
Many attempts were suggested in terms of the constitute of network architecture, including residual blocks~\cite{he2016deep}, squeeze-and-excitation~\cite{li2018recurrent}, and recurrent networks~\cite{ren2019progressive}.
Several methods were proposed to improve computational efficiency by designing lightweight networks in a cascade manner~\cite{fan2018residual} and a Laplacian pyramid framework~\cite{fu2019lightweight}.
In addition, some useful priors, such as multi-scale~\cite{yasarla2019uncertainty,jiang2020multi}, bi-level layer prior~\cite{mu2018learning}, wavelet transform~\cite{yang2019scale} and dictionary learning mechanism~\cite{wang2020model} were also embedded into the deep learning-based methods for representing rain streaks. 
Moreover, semi-supervised learning approaches~\cite{wei2019semi,yasarla2020syn2real,huang2021memory} were proposed to leverage synthetic and unpaired real rain data more effectively. Recent Transformation-based methods~\cite{valanarasu2021transweather,zamir2022restormer,tu2022maxim} also emerged in the domain of image de-raining. Nonetheless, the reliance on synthetic data for training still prevails in the aforementioned methods.

Taking a different approach, TimeLapsNet~\cite{cho2020single} introduced a de-raining network relying solely on a real rain dataset, where both the camera and scene remain static except for time-varying rain streaks. MemoryNet~\cite{cho2022memory} extended this concept~\cite{cho2020single} by incorporating an external memory network, albeit at the cost of additional memory usage. Regrettably, both methods encounter limitations in terms of practical real-world applications due to their demanding computational requirements. In contrast, our proposed method introduces a novel attention operator, making it more amenable to real-world scenarios.
\vspace{-5pt}
\subsection{Attention mechanism}
Attention mechanisms, designed to prioritize relevant regions of an image while filtering out irrelevant regions, have attracted significant attention in many research field. In the context of a vision system, an attention mechanism functions akin to a dynamic selection process, updating feature weights adaptively based on the significance of the input. Leveraging attention mechanisms, prototype networks are devised to generate embeddings that cluster multiple points around a central prototype representation for each class. This approach involves learning class-specific prototypes through feature space averaging. This concept has found wide application in computer vision tasks like few-shot classification~\cite{snell2017prototypical} and semantic segmentation~\cite{liu2020part}.

In recent developments, MOSS~\cite{huang2021memory} and MemoryNet~\cite{cho2022memory} introduced the utilization of memory networks to store prototypical rain streak-related features, albeit at the expense of substantial memory consumption. In contrast, our contribution involves the adoption of a prototype network with integrated attention mechanisms, eliminating the need for additional memory usage.


\begin{figure*}[!]
	\centering
	{\includegraphics[width=0.8\textwidth]{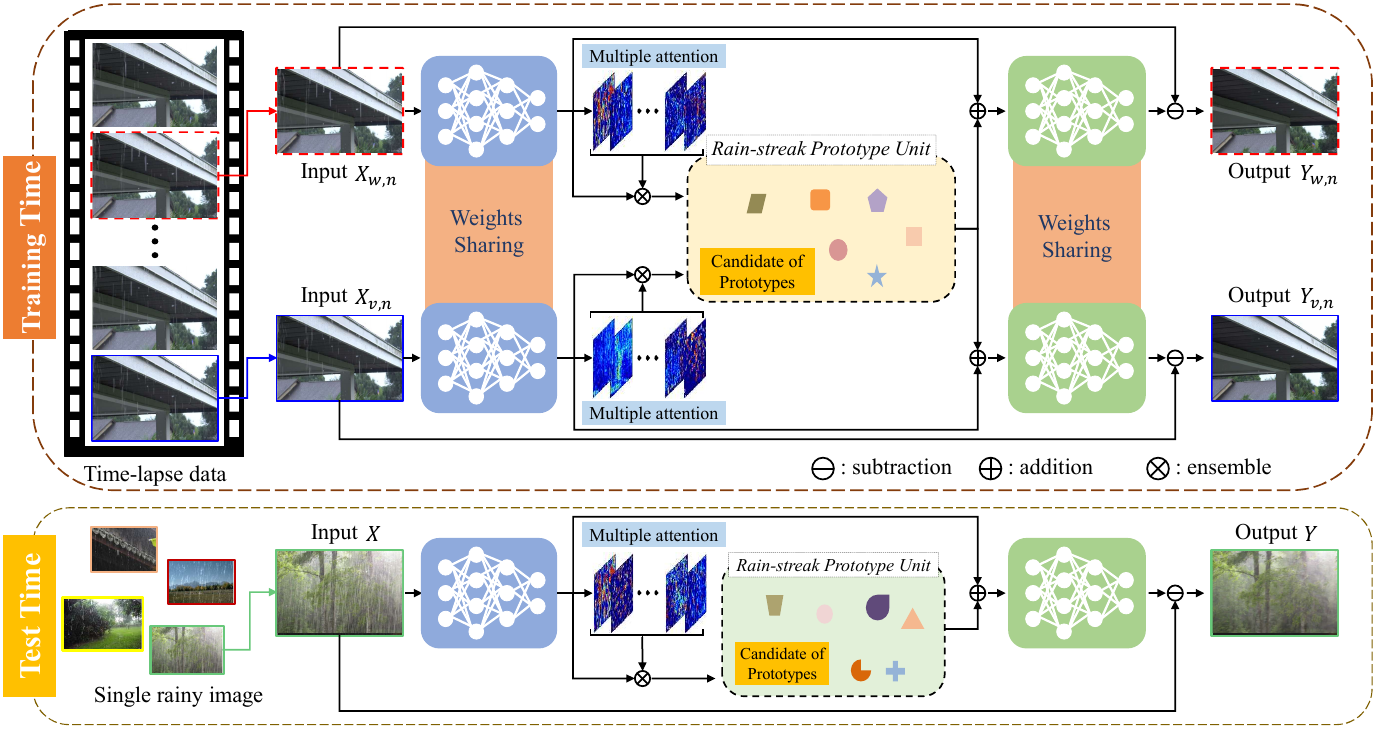}}
    \vspace{3pt}
	\caption{The overall framework of our method.}
	\label{fig:2}
\end{figure*}
\vspace{-5pt}

\section{Proposed Method}
	
We show an overview of our framework in Fig.~\ref{fig:2}.
For training, we use a set of time-lapse data $\mathbf{X} = \{X_{n}\}_{n=1, ...,N}$ and ${X}_{n}=\{{X}_{t,n}\}_{t=1, ...,T}$, where $N$ is the number of all time-lapse data, $T$ is the whole time of the time-lapse data.
$n$ and $t$ mean the index of scenes and the time, respectively.
The objective is to infer a de-rained image $Y_{t,n}$ for each $X_{t,n}$ through the proposed method. Note that, for simplicity, the framework repeats the same process for each $t$ and $n$ in the RsPU, where the subscripts of $t$ and $n$ are omitted.
\vspace{-10pt}

\subsection{Rain-streak Prototype Unit (RsPU)}

The Rain-streak Prototype Unit (RsPU) operates by learning and capturing rain-streak relevant features in the form of multiple prototypes. This concept bears resemblance to memory-guided methods~\cite{cho2022memory,huang2021memory}. While these methods calculate the similarity between queries and memory items using external networks to store rain streak-related features, they incur memory consumption. In contrast, our RsPU employs self-attention mechanisms to generate prototypical rain streak features internally, eliminating the need for extra memory usage.

We denote by $\mathbf{x}$ a corresponding feature of rain image ${X}$ from the de-raining network.
The de-raining network inputs the rainy image ${X}$ and gives the extracted features $\mathbf{x}$ of size $H \times W \times C$, where $H$, $W$, and $C$ are height, width, and the number of channels, respectively.
We denote by $\mathbf{x}^{k}\in\mathbb{R}^{C} (k = 1, ..., K)$, where $K = H \times W$, individual extracted features of size $1 \times 1 \times C$ in the extracted features $\mathbf{x}$.
An attention mapping functions\footnote{Attention mapping functions are implemented as fully connected layers to generate multiple attention maps and form a candidate of prototypes.} $\{\mathcal{A}_{m}:\mathbb{R}^{C}\rightarrow\mathbb{R}^{1}\}^{M}_{m=1}$ are employed to assign rain streak weights to encoding vectors, $w^{k,m}\in\mathcal{W}^{m}=\mathcal{A}_{m}(\mathbf{x})$.
On each pixel location, the rain streak weight measures the probability of finding rain streaks range of the {encoding vector.}
We denote by $\mathcal{W}^{m}\in\mathbb{R}^{H \times W \times 1}$, where the $m-$th rain streaks map generated from the $m-$th attention function.
Then one unique prototype $\mathbf{p}^{m}$ is derived as an aggregation of $K$ encoding vectors with normalized rain streak weights as:
    \begin{equation}~\label{eq:1}
    	\mathbf{p}^{m} = {\sum\limits_{k = 1}^K\frac{w^{k,m}}{\sum\nolimits_{k' = 1}^K {w^{k',m}}}}\mathbf{x}^{k}.
    \end{equation}	
Similarly, $M$ prototypes are derived from multiple attention functions to form a candidate of prototype, $\mathcal{P}=\{\mathbf{p}^{m}\}^{M}_{m=1}$.
    
Input encoding vectors $\mathbf{x}^{k}(k\in K)$ from the encoding map of the de-raining network are used as queries to retrieve rain streak relevant features in the candidate of prototype for estimating a rain streak encoding $\mathbf{\hat{X}}\in\mathbb{R}^{H \times W \times C}$. 
For every obtained rain streak encoding vector obtained by
    \begin{equation}\label{eq:2} 
    	\mathbf{\hat{x}}^{k} = {\sum\limits_{m = 1}^M\alpha^{k,m}{\mathbf{p}^{m}}},
    \end{equation}	

where $\alpha^{k,m}$ denotes the relevant score between the $k-$th encoding vector $\mathbf{{x}}^{k}$ and the $m-$th prototype item $\mathbf{p}^{m}$.
    
Using a channel-wise summation, we aggregate the obtained rain streak map and the original encoding $\mathbf{{x}}$ as the final output.
The output encoding of RsPU is inputted to the decoder for estimating the rain streak {$\hat{R}$.}
Note that we aim to enrich {encoded features} to enhance estimation for various real rain streak information while suppressing consistent background information in time-lapse data.

 \subsection{Loss Functions}
    
For learning our framework, as in previous works~\cite{cho2020single,cho2022memory}, we utilize several loss functions including background consistency, cross consistency, and self-consistency loss.
Moreover, we propose a novel feature prototype loss, enabling prototype learning for enhanced prototypical rain streak features.

\paragraph{\textbf{Feature Prototype Loss}}

The feature prototype loss is designed in a form that reduces intra-rain streak variations while enlarging the inter-rain streak differences simultaneously.
Contrary to the conventional prototype loss function’s optimization on positive and negative pairs, the proposed loss calculates gradients and does backpropagation based on the overall distance of prototypes within RsPU.
The feature prototype loss consists of two terms, including cohesion loss $\mathcal{L}_{coh}$ and divergence loss $\mathcal{L}_{div}$, formulated as:
\begin{equation}~\label{eq:6}
{{\mathcal{L}_{fea}}}={{\mathcal{L}}_{coh}}+\lambda_{a}{{\mathcal{L}}_{div}},
\end{equation}
where $\lambda_{a}$ is the weighting factor.
The cohesion loss encourages the rain streak relevant features to be gathered with compact prototypes.
It penalizes the mean $L_{2}$ distance between the extracted features $\mathbf{x}$ from the de-raining network and their most-relevant prototypes as:
\begin{equation}\label{eq:7}
{{\mathcal{L}}_{coh}}={\frac{1}{K}}\sum\limits_{k = 1}^K{{{\left\| \mathbf{{x}}^{k} -{\mathbf{p}^{*}} \right\|}_{2}}},
\end{equation}
\begin{equation}\label{eq:8}
\mathrm{s.t.,*} = \underset{m\in[1,M]}{\mathrm{argmax}}\alpha^{k,m},
\end{equation}
where $\alpha^{k,m}$\footnote{
{$\mathrm{argmax}$ is not involved in the back-propagation and is only used to obtain indices of the most relevant vector.}} is the relevant score.

To promote the diversity among prototype items by pushing the learned prototypes away from each other, we propose the divergence term $\mathcal{L}_{div}$ defined with a margin of $\delta$ as:
\begin{equation}\label{eq:9}
{{\mathcal{L}}_{div}}={\frac{1}{M(M-1)}}\sum\limits_{m = 1}^M\sum\limits_{m' = 1}^M{{ [-{\left\| {\mathbf{p}^{m}} - {\mathbf{p}^{m'}} \right\|}_{2}} + \delta ]_{+}}.
\end{equation}
With this feature prototype loss, the network captures the differentiated rain streak features in the RsPU. This loss is well-tailored to the RsPU in that the prototypes are encouraged to encode compact and diverse rain streak features.

\vspace{-5pt}

\paragraph{Background Consistency Loss}
This loss encourages the generation of consistent background images between the input images~\cite{cho2020single,cho2022memory}.
We use the following equation to compute the loss.
\begin{equation}\label{eq:}
	{{\mathcal{L}}_{b}}={\sum\limits_{n\in N} {\sum\limits_{\{w,v\}\in T}}} {\sum\limits_{i} {{{{\left\| {\hat{Y}_{w,n}(i)}-{\hat{Y}_{v,n}(i)} \right\|}_{1}}}}},	
\end{equation} 
where $\hat{Y}_{w,n}$ is a background image estimated from $X_{w,n}$ and ${\left\| \cdot \right\|}_{1}$ denotes the $L_{1}$ distance.
${w,v}$ represents the different times in $T$.
${\hat{Y}_{w,n}(i)}$ and ${\hat{Y}_{v,n}(i)}$ are the values at pixel $i$ from the image ${\hat{Y}_{w,n}}$ and ${\hat{Y}_{v,n}}$, respectively.

\paragraph{Cross Consistency Loss}
This loss aims to estimate the approximation of the overall structure of the layout information~\cite{lettry2018unsupervised,lettry2018deep} obtained by

\begin{equation}\label{eq:cross}
	{{\mathcal{L}}_{c}}={\sum\limits_{n\in N} {\sum\limits_{\{w,v\}\in T}}} {\sum\limits_{i}{{{{\left\| {X_{w,n}(i)}-{\hat{Y}_{v,n}(i)} \right\|}_{1}}}}}.
\end{equation}
This loss helps the network achieve good initial results in initial training.
	
\paragraph{Self Consistency Loss}
This loss makes the summation of the estimated $\hat{Y}$ and $\hat{R}$ to be the input image ${X}$~\cite{cho2020single,cho2022memory} obtained by
\begin{equation}\label{eq:self}
	{{\cal L}_s} = {\sum\limits_{n\in N} {\sum\limits_{w\in T}}} {\sum\limits_{i} {{\left\| {{X_{w,n}(i)} - ({\hat{Y}_{w,n}(i)} + {\hat{R}_{w,n}(i)}) }\right\|}{_1}}}.
\end{equation}
This loss acts like a regularize.

\paragraph{Total Loss}
During the entire training process of the proposed network, the total loss is formulated as follows:
\begin{equation}~\label{eq:11}
	{{\mathcal{L}}_{tot}}={{\mathcal{L}}_{b}}+\lambda_{c}{{\mathcal{L}}_{c}}+\lambda_{s}{{\mathcal{L}}_{s}}+\lambda_{f}{{\mathcal{L}}_{fea}},
\end{equation}
where $\lambda_{c}$, $\lambda_{s}$, and $\lambda_{f}$ are the hyper-parameters determined by empirical experiments.

  \subsection{Network Architectures}

Our de-raining network framework closely adheres to the widely utilized encoder--decoder architecture, a prevailing choice in the realm of single image de-raining~\cite{cho2020single,yasarla2020syn2real,zhang2018density,cho2022memory}. In this paradigm, all convolutional layers employ a $3\times3$ kernel size. The encoder employs max-pooling layers with $2\times2$ kernel size and stride, effectively reducing feature dimensions by a factor of 2. The input to the encoder is a rain image, denoted as ${X}$.
   
The decoder inputs the retrieved prototypical rain streak features from the RsPU and encoded features $\mathbf{x}$ to produce initial rain streak information.
In the decoder, each layer comprises $3 \times 3$ deconvolution and convolution layers followed by ReLU, which is connected to the encoder using skip connections.
The deconvolution layer, implemented with transposed convolutional layers, has an upscaling factor of 2.
As a result, the de-rained image is obtained by subtracting the estimated rain streaks from the input rain image.
The details of network architectures are described in the supplementary material.

\section{Experiment}

A demonstration is given for the proposed method against several state-of-the-art image de-raining methods including JORDER-E~\cite{yang2019joint}, DDN~\cite{fu2017removing}, PReNet~\cite{ren2019progressive}, SPANet~\cite{wang2019spatial}, RCDNet~\cite{wang2020model}, MPRNet~\cite{zamir2021multi}, SIRR~\cite{wei2019semi}, NLEDN~\cite{li2018non}, and MOSS~\cite{huang2021memory}. 
Additionally, the performances are compared to the proposed method with state-of-the-art Vision Transformer-based methods such as Restormer~\cite{zamir2022restormer}, MAXIM~\cite{tu2022maxim}, and DRSformer~\cite{chen2023learning} and time-lapse data-based methods including TimLapseNet~\cite{cho2020single} and MemoryNet~\cite{cho2022memory}.
We used publicly sourced codes and pre-trained models provided by the authors to produce the de-raining results.

\paragraph{Datasets}

We used a time-lapse benchmark provided by TimLapseNet~\cite{cho2020single} to train the proposed method. 
They provide rain image pairs comprised of 2 images sampled from 30 images from the 186 total scenes. 
For a fair comparison, the identical experimental setups were used in the previous works~\cite{cho2020single,cho2022memory}.
Since our framework aims to the consistent background information between time-lapse data, we use only time-lapse data during training without ground truth.
    
For testing, we conducted experiments on real and synthetic datasets, respectively. 
For the real dataset, we used the \textbf{RealDataset} to provide real rain images with realistic ground-truth background images generated by a semi-automatic algorithm~\cite{wang2019spatial}, thus enabling the quantitative evaluations.
We also obtained real-world rain images from the Internet and previous studies~\cite{zhang2018density,wang2019spatial,yang2019joint}.
For the synthetic dataset, following ~\cite{cho2022memory,cho2020single}, we used \textbf{Rain14000}~\cite{fu2017removing}, \textbf{Rain12000}~\cite{zhang2018density}, and \textbf{Rain100}~\cite{yang2019joint}.

\paragraph{Metrics}
Following a de-raining works~\cite{yang2020single}, we adopted two metrics for quantitative evaluation of performance by using the paired synthetic data: peak signal-to-noise ratio (PSNR) and structural similarity index measure (SSIM).

\paragraph{Implementation Details}
The proposed networks were trained using the PyTorch library with an Nvidia RTX TITAN GPU.
{As described in the TimLapseNet~\cite{cho2020single} and MemoryNet~\cite{cho2022memory}, 
we did not utilize any additional data augmentation, such as horizontal and vertical flips, due to large-scale real rain data having been provided. The input images were resized to the resolution of 256 $\times$ 256 and normalized to the range of [-1, 1]. During the training, the learning rate was fixed as $10^{-4}$ and set the batch size as 16.
We set the height $H$ and width $W$ of the extracted features $\mathbf{x}$ feature map, the number of feature channels $C$, and candidates of rain streak maps $M$ to 256, 256, 128, and, 20, respectively.
For the feature prototype loss, $\lambda_{a}$ and $\delta$ are set to $0.1$ and $1$, respectively.
For the total loss hyper-parameters, we empirically set as $\lambda_{c}$ = 0.1, $\lambda_{s}$ = 0.001, and $\lambda_{f}$ = 0.1. 

\begin{figure}[!t]
	\centering
	\subfigure
	{\includegraphics[width=0.115\textwidth]{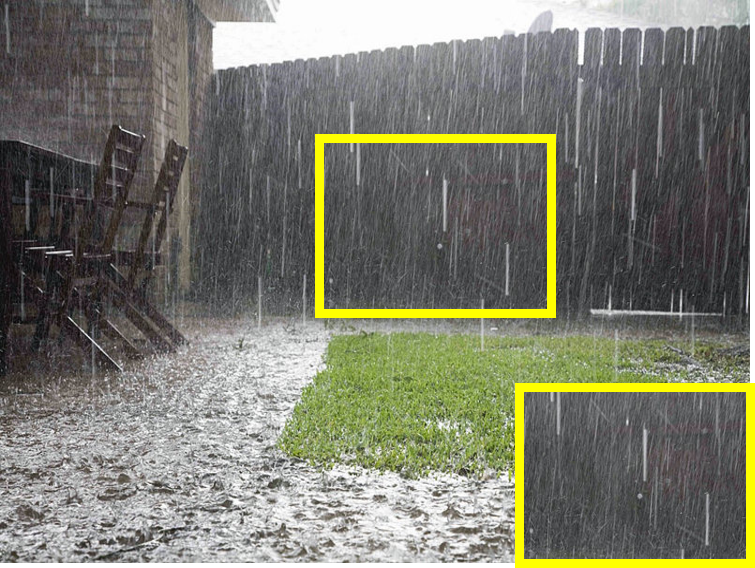}}
    \subfigure
	{\includegraphics[width=0.115\textwidth]{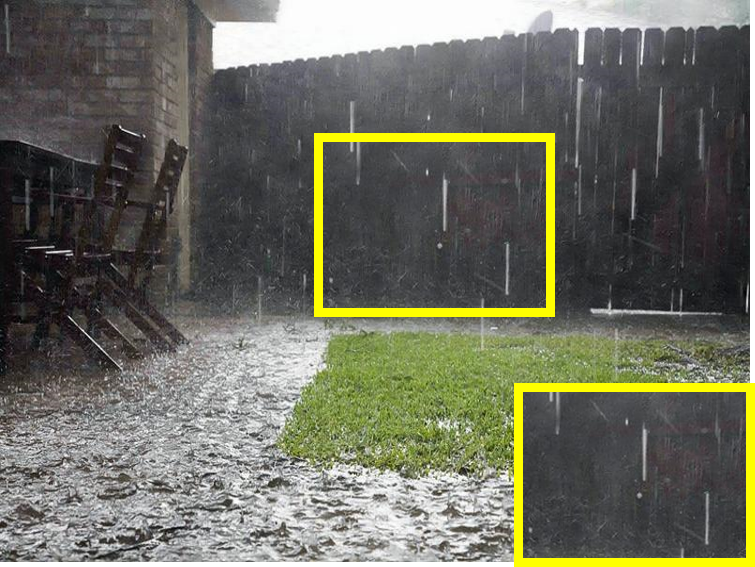}}
    \subfigure
	{\includegraphics[width=0.115\textwidth]{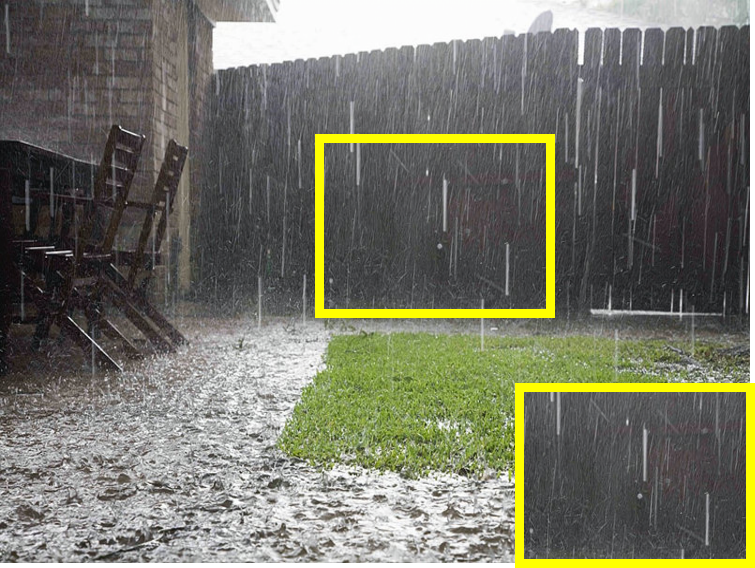}}
    \subfigure
	{\includegraphics[width=0.115\textwidth]{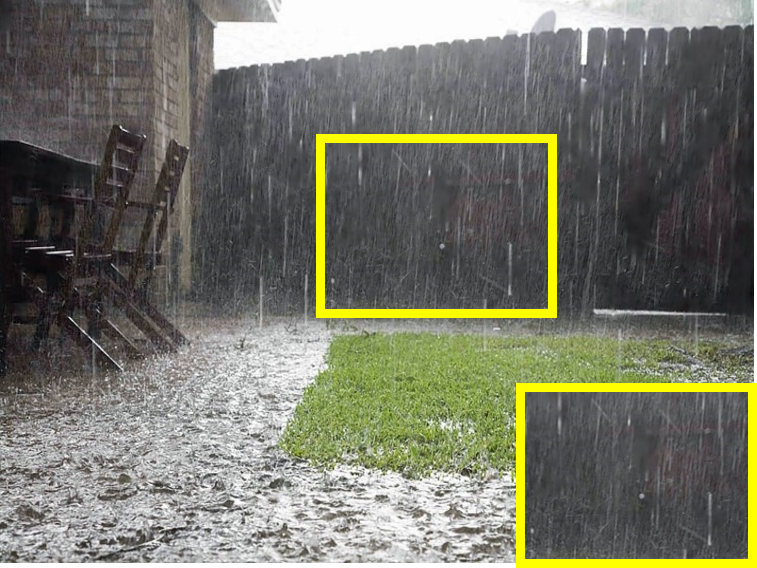}}
	\subfigure
	{\includegraphics[width=0.115\textwidth]{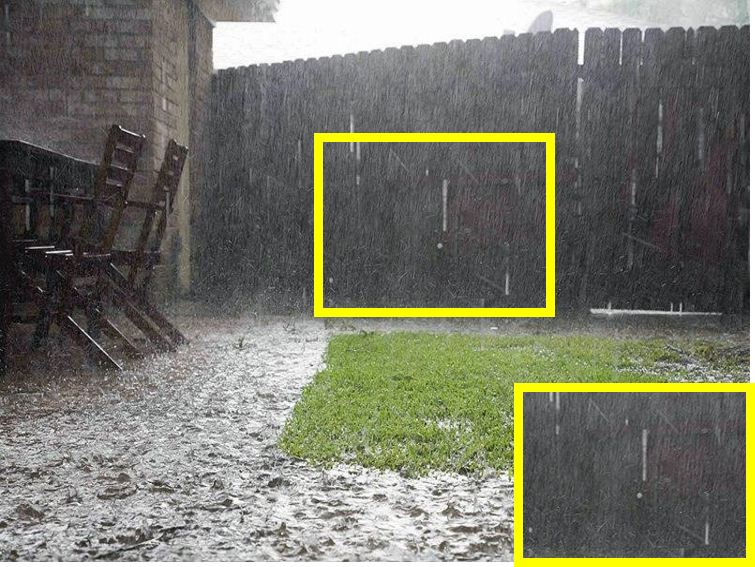}}
    \subfigure
	{\includegraphics[width=0.115\textwidth]{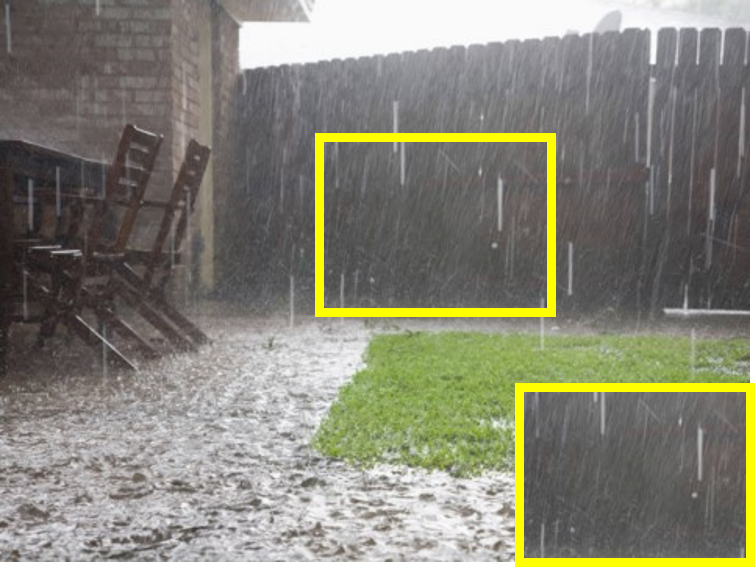}}
    \subfigure
	{\includegraphics[width=0.115\textwidth]{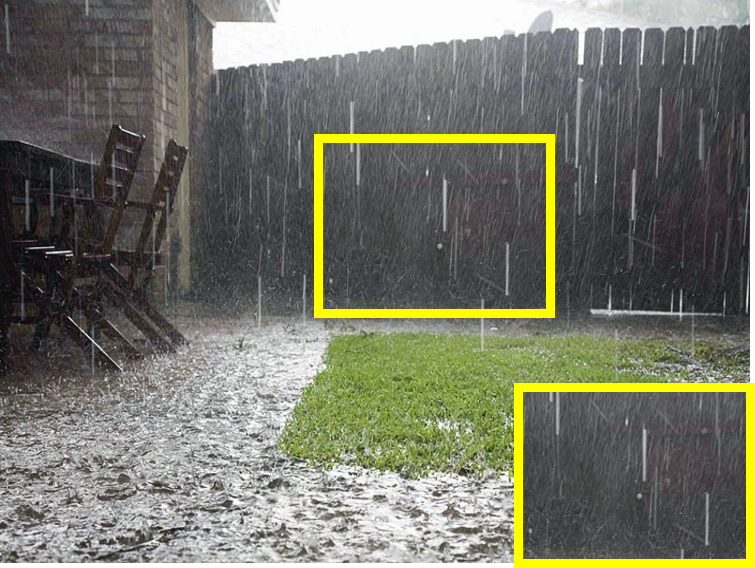}}
    \subfigure
    {\includegraphics[width=0.115\textwidth]{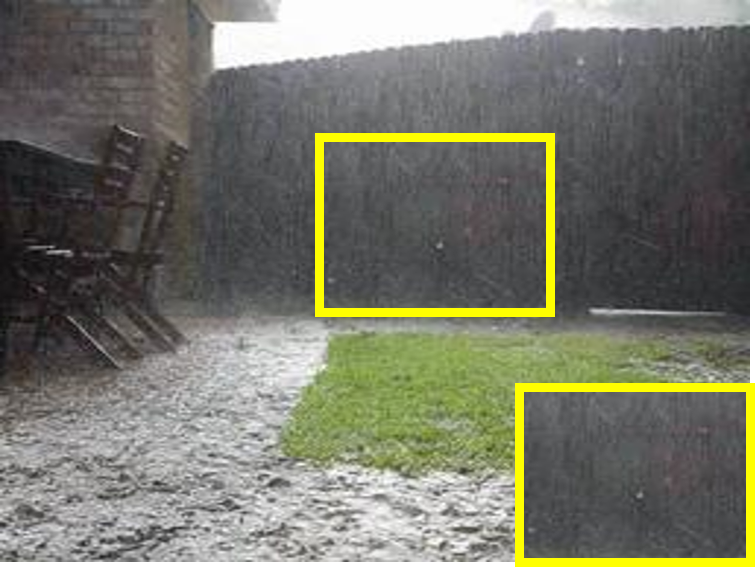}}
	\\
	\vspace{-8pt}
    \subfigure
	{\includegraphics[width=0.115\textwidth]{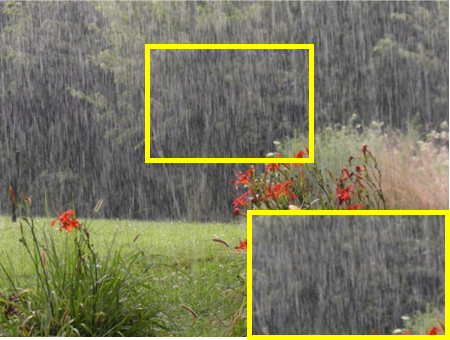}}
    \subfigure
	{\includegraphics[width=0.115\textwidth]{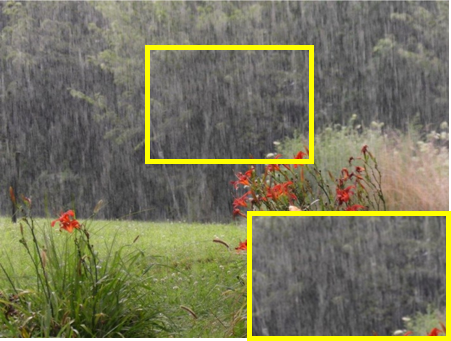}}
	\subfigure
	{\includegraphics[width=0.115\textwidth]{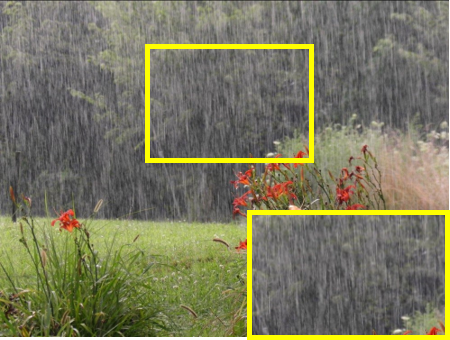}}
    \subfigure
    {\includegraphics[width=0.115\textwidth]{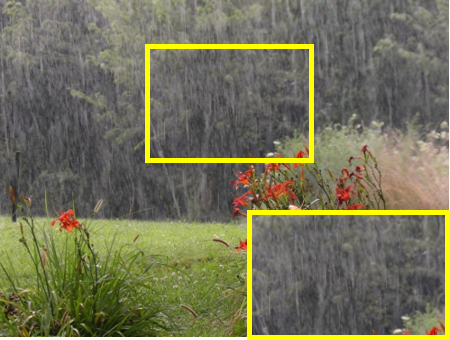}}
	\subfigure
	{\includegraphics[width=0.115\textwidth]{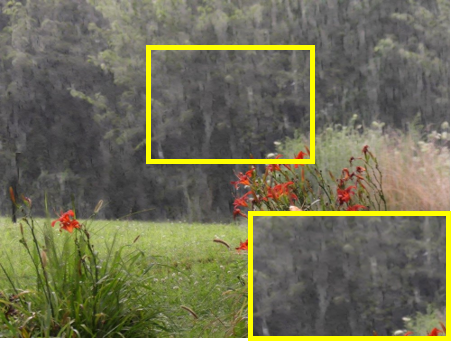}}
    \subfigure
	{\includegraphics[width=0.115\textwidth]{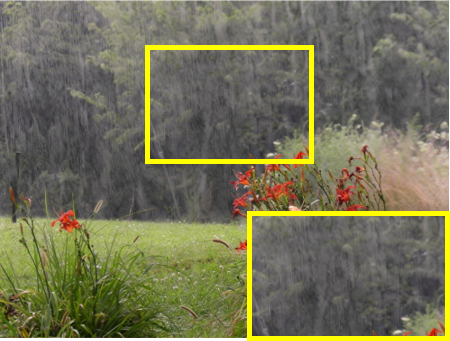}}
	\subfigure
	{\includegraphics[width=0.115\textwidth]{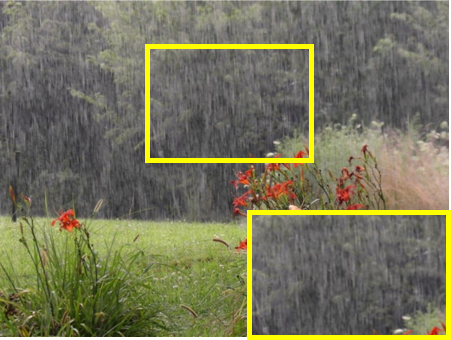}}
    \subfigure
	{\includegraphics[width=0.115\textwidth]{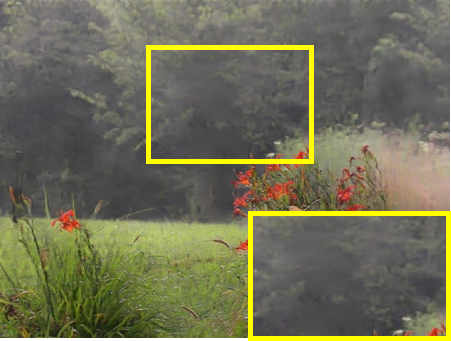}}\\
	\vspace{5pt}
	\caption{Qualitative results on real rain images. From left to right, Input image, JORDER-E, MPRNet, RCDNet, MemoryNet, MAXIM, Restormer and Ours.}\label{fig:3}\vspace{-1pt}
\end{figure}
\begin{figure}[!t]
	\centering
	\subfigure[]
	{\includegraphics[width=0.115\textwidth]{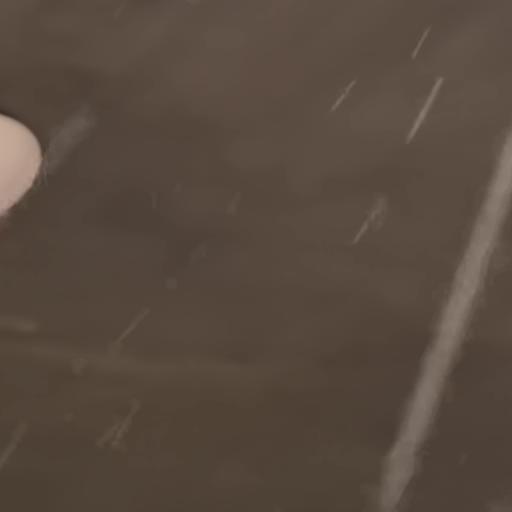}}
    \subfigure[]
	{\includegraphics[width=0.115\textwidth]{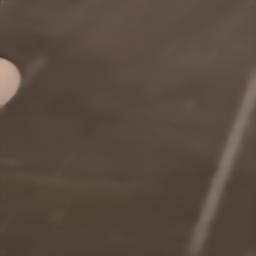}}
    \subfigure[]
	{\includegraphics[width=0.115\textwidth]{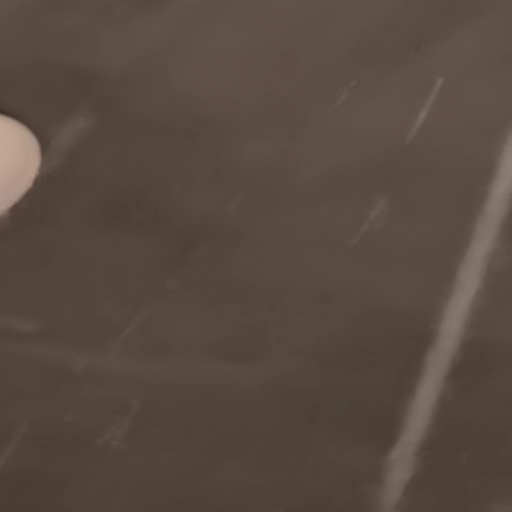}}
    \subfigure[]
	{\includegraphics[width=0.115\textwidth]{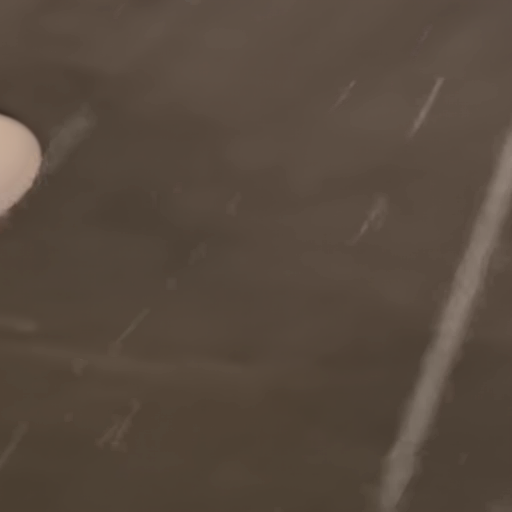}}
	\subfigure[]
	{\includegraphics[width=0.115\textwidth]{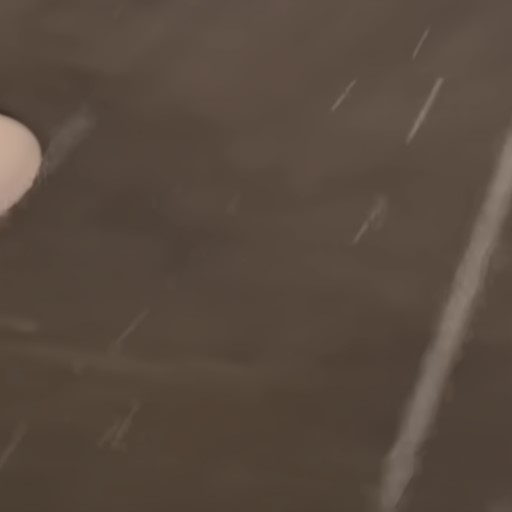}}
    \subfigure[]
	{\includegraphics[width=0.115\textwidth]{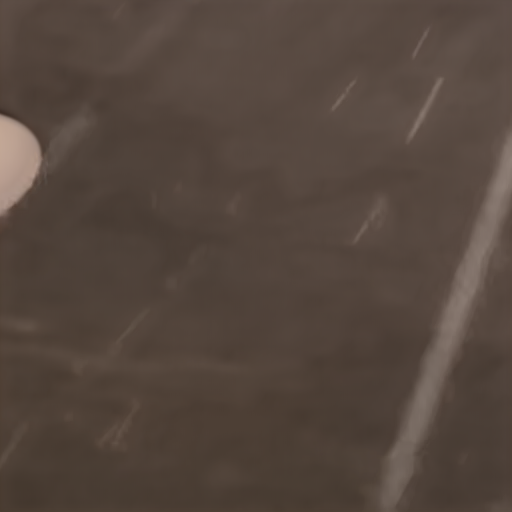}}
    \subfigure[]
	{\includegraphics[width=0.115\textwidth]{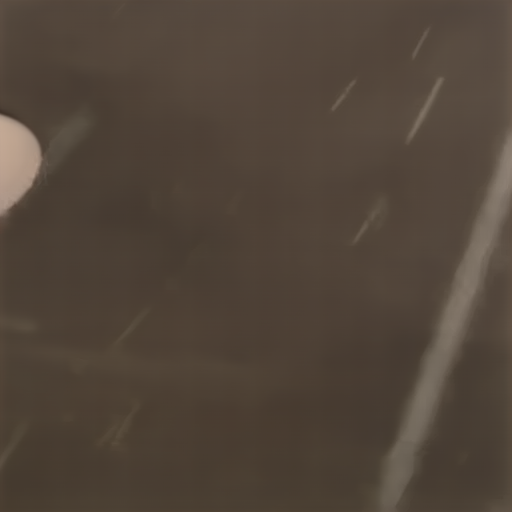}}
	\subfigure[]
	{\includegraphics[width=0.115\textwidth]{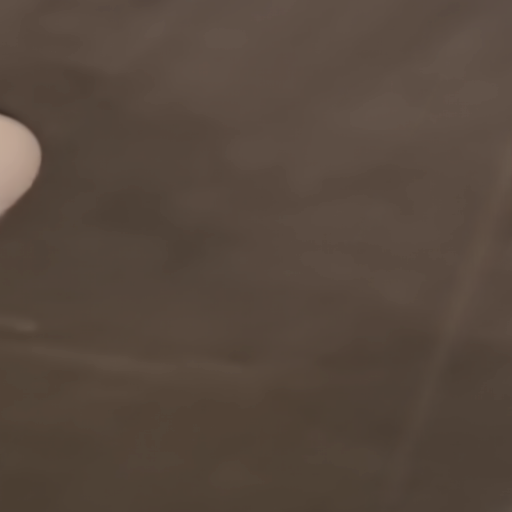}}\\
	\vspace{5pt}
	\caption{{Qualitative results on RealDataset}~\cite{wang2019spatial}. (a) Input, (b) MemoryNet, (c) JORDER-E, (d) RCDNet, (e) Restormer, (f) MAXIM, (g) DRSformer, and (h) ours.
    } \label{fig:4}\vspace{-10pt}
\end{figure}
\subsection{Comparison with State-of-the-Art} 
 \paragraph{Real Rain Analysis}
For evaluating the generalization ability of real rain images, we conducted a qualitative evaluation compared to all competing state-of-the-art methods.
In the first row of Fig.~\ref{fig:3}, while existing methods suffer from removing long and thin real rain streaks, our method effectively handles various types of real rain streaks.
The second row shows that our method outperforms the competing method in terms of discriminating the rain streaks and background information well even in the rain accompanied by fog.
This is because existing methods heavily rely on synthetic data, which could not cover various types of real rain streaks.
Additionally, Fig.~\ref{fig:4} shows de-rained results on RealDataset~\cite{wang2019spatial}. 
In contrast, our method confidently removes rain streaks compared to other methods thanks to the advantage of the RsPU and real-world time-lapse data. Additional qualitative results are shown in the supplementary material.

\paragraph{Synthetic Rain Analysis}
    
Table~\ref{tab:1} shows the quantitative results of recent studies, from CNN-based methods to transformer-based methods, on various benchmarks. 
Although our method does not use any ground-truth data for training, we achieve comparable performance gain and even outperform compared to state-of-the-art methods.
This means leveraging a real rain dataset effectively extracts useful rain streak features, even in the synthetic rain dataset.
We believe that our framework, which intelligently combines the real dataset and RsPU, effectively estimates the prototypical rain streak features.

 \begin{table*}
				\centering
				\renewcommand{\tabcolsep}{3mm}
    \scalebox{0.71}
    {
				\begin{tabular}{l|c|c|cc|cc|cc|cc}
					\toprule 
					\multirow{3}{*}{Benchmark} &  \multirow{3}{*}{GT} &  \multirow{3}{*}{T-L} & \multicolumn{6}{c|}{Synthetic Dataset}  & \multicolumn{2}{c}{Real-world Dataset} \\ \cline{4-11}
						&  &  & \multicolumn{2}{c|}{\textbf{Rain14000}}  & \multicolumn{2}{c|}{\textbf{Rain12000}} & \multicolumn{2}{c|}{\textbf{Rain100}}  & \multicolumn{2}{c}{{\textbf{RealDataset}}} \\ \cline{4-11}
					&   &  &  PSNR  & SSIM  &  PSNR  & SSIM  &   PSNR  & SSIM    &  PSNR & SSIM \\ 
					\midrule
					DDN~\cite{fu2017removing}  & Yes    & No         & 28.45 & 0.889 & 30.97 & 0.912 &   34.68   & 0.967   &  36.16   & 0.946   \\
					DID~\cite{zhang2018density}& Yes   & No      & {26.17} & {0.887} &  31.30 & 0.921 & 35.40 &  0.962   &  28.96    & 0.941    \\
					MPRNet~\cite{zamir2021multi}& Yes     & No      & 33.64  & 0.938 & 32.91 &  0.916 & 36.40  &  0.965   &  40.12    &  0.984   \\
					PReNet~\cite{ren2019progressive}& Yes  & No       & {{32.55}} & {\textbf{{0.946}}} &  33.17 & 0.942 &   37.80   & {{0.981}}  &  40.16    &  0.982  \\
					SIRR~\cite{wei2019semi}& Yes   & No      & 28.44 & 0.889 & 30.57 & 0.910 &   34.75   & 0.969    &  35.31    &  0.941  \\
					JORDER-E~\cite{yang2019joint} & Yes   & No      & 32.00 & {{{0.935}}} & {33.98} & {\underline{0.950}} &   {{38.59}}  & {\underline{0.983}}     &  {{40.78}}     &  0.981   \\
					NLEDN~\cite{li2018non}& Yes     & No      & {29.79} & {0.897} & {33.16} & {0.919} & 36.57 &  0.975   &  40.12    &  0.984    \\
					RCDNet~\cite{wang2020model}  & Yes  & No  & 30.66 & 0.921 &  31.99 & 0.921 &  {\textbf{40.17}}  & {\textbf{0.988}}  &  {{41.47}}   &  {{0.983}}  \\
					{MOSS~\cite{huang2021memory} } & Yes  & No  & 31.22 & 0.932 &  32.87 & 0.932 &  {{37.67}}  & {{0.974}}  &  {{40.01}}   &  {{0.971}}  \\
     					SPANet~\cite{wang2019spatial}& Yes   & Yes      & {29.85} & {0.912} &  33.04 & {{0.949}} &   35.79   & 0.965  &  40.24    &  0.981   \\
     {Restormer~\cite{zamir2022restormer}} & Yes  & No  & \underline{34.18} & 0.944 & 33.19  & 0.926 &  38.99  & 0.978  &   41.12  & 0.985    \\
     {MAXIM~\cite{tu2022maxim} } & Yes  & No  & 33.80 & 0.943 &  32.37 & 0.922 &  38.06  & 0.977  &  39.15   & 0.978   \\
     					\midrule 
					TimeLapsNet~\cite{cho2020single}  & No  & Yes  & {{{33.73}}} & 0.941 &  {{33.25}} & 0.935 &    37.89  &{{0.980}}   &  38.54    &   {\underline{0.989}}   \\
     MemoryNet~\cite{cho2022memory}  & No & Yes  & {{34.02}} & {\underline{0.945}} & {\underline{34.55}}& {{{0.949}}} &  {{38.45}}& {{0.981}}  &  {\underline{41.56}}    & \underline{{0.989}} \\
					\midrule
					Ours  & No & Yes  & {\textbf{34.53}} & \underline{{0.945}} & {\textbf{35.25}}& \textbf{0.951} &  {\underline{39.75}}& {{0.981}}  &  {\textbf{42.16}}    & \textbf{{0.990}} \\
					\bottomrule
		        \end{tabular}}    
    \vspace{10pt}
    \caption{
				{Quantitative comparison of single image de-raining using synthetic and {RealDataset}.} 
					GT and T-L denote using paired ground truth and time-lapse data, respectively.
					The best result is shown in bold, and the second--best is underlined. 
					}\label{tab:1}
			\end{table*}

\paragraph{Analysis of Run Time and Parameters}
\begin{table*}[t!]
	\centering
	\renewcommand{\tabcolsep}{3mm}
	\scalebox{0.6}{   
		\begin{tabular}[b]{c|ccccccccc}
			\toprule
			
		    & NLEDN & JORDER-E & RCDNet & MAXIM & Restormer & {DRSformer} & TimLapseNet & MemoryNet & Ours \\\cline{1-10}
			\midrule 
			GPU &  1.17 & 1.74  & 0.85 & 4.12 & 6.32 & 6.58 & 3.87 & 0.82 & 0.52  \\
			Params. &  1.01 M & 4.17 M   & 3.17 M  & 14.1 M  & 26.12 M &   33.7 M & 10.2 M & 0.81 M & 0.61 M \\
			\bottomrule
		\end{tabular}
	}
    \vspace{5pt}
 	\caption{Comparison of run time (s) and the number of parameters.}
	\label{tab:4}
\end{table*}

We compare the running time and the number of parameters in Table~\ref{tab:4}.
To evaluate the running time, we used 100 images with a size of 1000 $\times$ 1000.
It is clear that our method has a comparable GPU runtime compared with other CNN-based methods~\cite{li2018non,yang2019joint} and is significantly faster than several Transformer-based methods~\cite{chen2023learning,zamir2022restormer,tu2022maxim}. This shows that our network can extract more effective representations, which could be executed on devices with limited computing power and memory in practice. In addition, Our method shows faster than MemoryNet~\cite{cho2022memory} requiring the external memory network, thanks to the proposed RsPU that does not require additional memory.

\subsection{Ablation Study}

 \paragraph{Study of the feature prototype loss}
To verify the effectiveness of the feature prototype loss ${\mathcal{L}_{fea}}$, we conducted the experiment according to the loss functions.
As described in previous studies~\cite{cho2020single,cho2022memory}, we show that the performance of experimental results is improved with the addition of loss.
For quantitative evaluation, Table~\ref{tab:3} (Left) shows the performance of our model trained with the proposed loss functions on RealDataset. The model trained with a ${\mathcal{L}_{fea}}$ achieves better results than the model trained without the ${\mathcal{L}_{fea}}$. The experimental results demonstrate the advantage of the proposed loss.
    
We show the de-rained image and the result of the estimated rain streaks learned with and without ${\mathcal{L}_{fea}}$ in Fig.~\ref{fig:5}.
Our model trained without ${\mathcal{L}_{fea}}$ is not effective at discriminating rain streaks and background, as shown in Fig.~\ref{fig:5} (b).
We achieved the improved performance with ${\mathcal{L}_{coh}}$ but it still shows the rain streaks in the face of the spider man in Fig.~\ref{fig:5} (c).
Our model trained with ${\mathcal{L}_{fea}}$ effectively discriminates rain streak and background in Fig.~\ref{fig:5} (d).
This demonstrates that ${\mathcal{L}_{fea}}$ helps to discriminate rain streaks effectively and yields improved de-raining performance.

\begin{figure}[!t]
	\centering
	\subfigure[Input]
	{\includegraphics[width=0.24\textwidth,height=0.09\textheight]{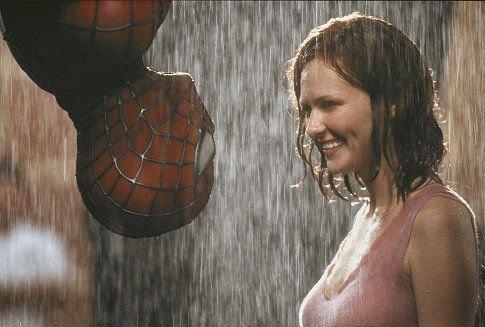}}
    \subfigure[$\mathcal{L}_{b}+\mathcal{L}_{c}+\mathcal{L}_{s}$]
	{\includegraphics[width=0.24\textwidth,height=0.09\textheight]{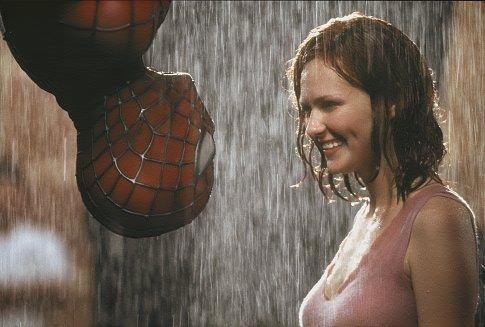}}
    \subfigure[$\mathcal{L}_{b}+\mathcal{L}_{c}+\mathcal{L}_{s}+\mathcal{L}_{coh}$]
	{\includegraphics[width=0.24\textwidth,height=0.09\textheight]{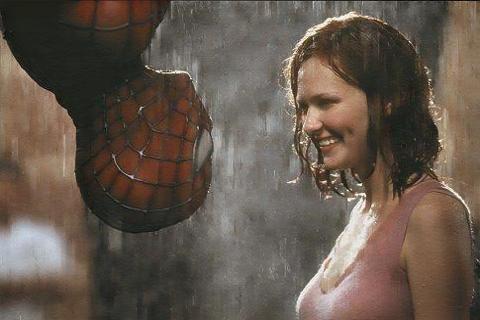}}
    \subfigure[$\mathcal{L}_{b}+\mathcal{L}_{c}+\mathcal{L}_{s}+\mathcal{L}_{fea}$]
	{\includegraphics[width=0.24\textwidth,height=0.09\textheight]{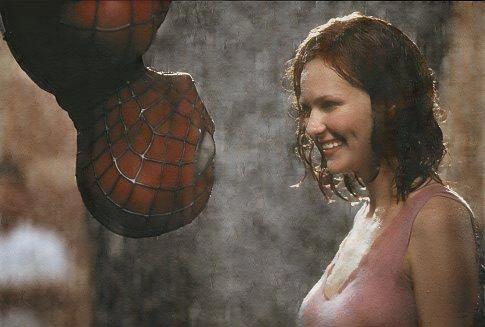}}
	\vspace{3pt}
	\caption{{Visualization results of de-raining results, in terms of the loss functions.}
    } \label{fig:5}\vspace{-10pt}
\end{figure}

\begin{table}[htbp!]
		\renewcommand{\tabcolsep}{4mm}
		\scalebox{0.75}{
			\begin{tabular}[b]{cccc|cc}
			\toprule
			$\mathcal{L}_{b}$  & $\mathcal{L}_{c}$ & $\mathcal{L}_{s}$  & $\mathcal{L}_{fea}$ & PSNR  & SSIM \\  \cline{1-6}
			\midrule					
			\cmark     &    \cmark   &   &       & 39.92 & 0.981 \\		
			\cmark     &    \cmark   &   &   \cmark    & 40.97 & 0.982 \\		
			&    \cmark  &  \cmark   &     & 40.81 & 0.982 \\
			&    \cmark  &  \cmark  &  \cmark  & 41.53 & 0.985 \\	
			\cmark     &    \cmark   &  \cmark  &     & {41.82} & 0.987 \\
			\cmark     &    \cmark    &  \cmark  & \cmark   & 42.16 & 0.990 \\
			\bottomrule
		\end{tabular}}
\hfill
    \renewcommand{\tabcolsep}{4mm}
		\scalebox{0.85}{
			\begin{tabular}[b]{c|ccc}
			\toprule
		& 	$\mathcal{A}$  & $\mathcal{B}$ & $\mathcal{C}$     \\   \cline{1-4}
			\midrule
			 De-rainingNet   &  \cmark  & \cmark   & \cmark       \\
			 Siamese   &     &  \cmark  &   \cmark    \\
			RsPU   &     &   &  \cmark   \\
			\hline
			PSNR   &  39.82   & 40.53   &    42.16    \\
			SSIM   &  0.980  &  0.983  &   0.990    \\
			\bottomrule
		\end{tabular}
  }
   \vspace{-10pt}
\caption{(Left) Ablation study of loss functions. (Right) Ablation study of the network architecture.}
\label{tab:3}
\end{table}
\vspace{-15pt}

\begin{figure*}
	\centering
	{\includegraphics[width=\textwidth]{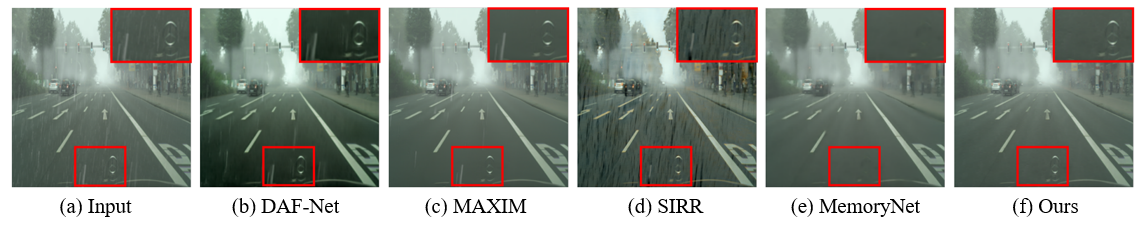}}
    \vspace{-8pt}
	\caption{{Comparison results from the proposed method against those from the state-of-the-art methods on real rain images accompanied with haze.}}
	\label{fig:6}\vspace{-13pt}
\end{figure*}

\paragraph{Study of Network Architectures}

To validate the effectiveness of our network, we evaluated the different combinations on \textbf{RealDataset}, as shown in Table~\ref{tab:3} (Right).
Compared to the primary de-raining network, with Siamese, our model with RsPU yields significantly improved de-raining performance.
This verifies that the self-attention mechanism is more effective at extracting informative rain streak features and estimating de-raining images than the static convolutional layer.

\section{Other weather condition}
We conducted experiments on heavy rain images often accompanied by haze effects using RainCityscapes.
DAF-Net proposed a rain imaging model with rain streaks and haze.
Although DAF-Net removes haze, they still remain rain streaks.
We applied our network to estimate de-rained images and compared them with SOTA. Fig.~\ref{fig:6} shows that existing methods tend to leave rain streaks.
SIRR generates a corrupted background. 
MemoryNet fails to preserve background information (\textit{i.e.}, the logo of the car is erased). While existing methods fail to remove the rain streaks and generate over-smoothing results, our method outperforms them even in haze.

\section{Conclusion} 
In summary, our paper introduces the Rain-streak Prototype Unit (RsPU) for efficient single-image de-raining. The RsPU utilizes an attention-based approach to encode diverse rain streak features as compact prototypes, overcoming memory limitations while capturing real rain complexities. Additionally, our proposed feature prototype loss enhances the discriminative power of these prototypes through a combination of cohesion and divergence components. Extensive evaluations demonstrate that our method achieves superior performance compared to existing techniques, highlighting its potential to advance rain removal capabilities for real-world applications.

\section*{Acknowledgement}

This work was supported by the National Research Foundation of Korea(NRF) grant funded by the Korea government(MSIP) (NRF-2021R1C1C2005202). (Corresponding author: Sunok Kim)

\bibliography{egbib}
\end{document}